\documentclass{llncs}

\usepackage{graphicx}
\usepackage{cite}
\usepackage{algorithm}  
\usepackage{algorithmicx}  
\usepackage{algpseudocode}  
\usepackage{amsmath}
\usepackage{subfig}
\usepackage{multirow}

\begin{document}
\title{Towards More Efficient Federated Learning with Better Optimization Objects\thanks
{This work was supported by NSFC under Grant No. 61936011 and No. 61521002, Beijing Key Lab of Networked Multimedia.}}

\author{Zirui Zhu\inst{1} \and
Ziyi Ye\inst{1}}

\authorrunning{Zhu et al.}

\institute{Dept. of Computer Science and Technology, Tsinghua University Beijing, China
\email{zhu-zr20@mails.tsinghua.edu.cn}\\
\email{yeziyi1998@gmail.com}}

\maketitle

\begin{abstract}
Federated Learning (FL) is a privacy-protected machine learning paradigm that allows model to be trained directly at the edge without uploading data. One of the biggest challenges faced by FL in practical applications is the heterogeneity of edge node data, which will slow down the convergence speed and degrade the performance of the model. For the above problems, a representative solution is to add additional constraints in the local training, such as FedProx, FedCurv and FedCL. However, the above algorithms still have room for improvement. We propose to use the aggregation of all models obtained in the past as new constraint target to further improve the performance of such algorithms. Experiments on various settings demonstrate that our method significantly improve the convergence speed and performance of the model.

\end{abstract}

\section{Introduction}

With the rapid development of technology and economy in recent years, more and more various types of intelligent devices have appeared in people’s lives, such as smartphones, wearable devices and IoT devices. When these devices are in use, a large amount of edge data is generated. If these data can be used for machine learning, it will significantly improve the user experience of intelligent devices. However, because this data is privacy-related and the amount is very large, the method of uploading this data to the data center for training is not feasible. In order to solve this problem, the learning paradigm called Federated Learning (FL) \cite{mcmahan2017communication, konevcny2015federated, konevcny2016federated} is proposed. FL allows the model to be trained directly at the edge nodes, and the knowledge learned by each node is summarized through the model aggregation process.

The most basic and commonly used algorithm in FL is Federated Averaging (FedAvg) \cite{mcmahan2017communication}. One of the biggest challenges faced by FedAvg is the heterogeneity of data at edge nodes. Studies \cite{mcmahan2016federated, zhao2018federated} have shown that when the data at edge nodes is non-iid, the convergence speed of FedAvg and the performance of the model will all decline. This performance degradation is considered to be related to gradient divergence caused by heterogeneous edge data. Therefore, an intuitive way to solve this problem is to add additional constraints in local training to reduce gradient divergence. According to the above ideas, some work has been done, and the most representative algorithms are FedProx \cite{FedProx}, FedCurv \cite{FedCurv} and FedCL \cite{FedCL}. FedProx is to directly constraint the changes of the local model compared with the global model in last round of training. FedCurv is to use the Fisher information matrix to impose greater penalties on the changes of important parameters. And FedCL is an efficient implementation method of FedCurv.

We can easily notice that the above algorithms all simply select the model obtained in the last round of training as the constraint target. So we try to design a new constraint target to optimize the performance of the above algorithms. Concretely, we refer to the idea of Temporal Ensembling \cite{TE2017} in semi-supervised learning, and use the aggregation of all models obtained in the past as new constraint target. Experiments on various settings demonstrate that new constraint target significantly improve the convergence speed of the algorithms and performance of the model.

\section{Related work}

\subsection{Federated Learning}

FL allows multiple nodes to train the model together without exposing data privacy. The most representative algorithm in FL is FedAvg \cite{mcmahan2017communication}, which mainly has the following steps:

\begin{itemize}
  \item[1)] Randomly select a certain number of nodes and distribute the global model to these nodes.
  \item[2)] Each node uses local data to train the model.
  \item[3)] The edge nodes upload the trained model, and the center performs model aggregation to obtain a new global model.
  \item[4)] Repeat steps 1-3.
\end{itemize}

More details are shown in Algorithm \ref{algorithm1}. In practical applications, Fedavg will face many challenges, the biggest of which is the heterogeneity of edge data. Heterogeneous edge data will lead to gradient divergence in local training, which will degrade the performance of Fedavg \cite{zhao2018federated}. In order to solve the problem caused by gradient divergence, some work has been done, among which the following three papers are the most representative.

The first one is FedProx \cite{FedProx}, which adds proximal term into the loss function of local training to limit the change of local model:

\begin{equation}
  \widetilde L(\omega) = L(\omega) + 
  \alpha \left \| \omega - \omega_{g} \right \|^2
\end{equation}

where $L(\omega)$ is the original local loss function, $\omega$ is the weight of local model, $\omega_{g}$ is the weight of global model obtained from the last model aggregation and $\alpha$ is the hyperparameter that controls the weight of the proximal term.

The second one is FedCurv \cite{FedCurv}, which use Fisher information matrix to evaluate the importance of each parameters in the model and impose greater penalties on changes of important parameters. The loss function is as follows:

\begin{equation}
  \widetilde L_{s}(\omega) = L_{s}(\omega) + \alpha \sum_{j \in S \backslash s}
  (\omega - \hat \omega_{j})^{T} diag(\hat I_{j}) (\omega - \hat \omega_{j})
\end{equation}

where $L_{s}(\omega)$ is the original local loss function, $S$ are nodes participating in this round of training, $s$ is current node, $\hat \omega_{j}$ is the model obtained from the last round of training on node j, $\hat I_{j}$ is the Fisher information matrix corresponding to $\hat \omega_{j}$ and $\alpha$ is the hyperparameter that controls the weight of the penalty term.

The third one is FedCL \cite{FedCL} which is an efficient implementation method of FedCurv, it keeps a proxy dataset at the central node so that the calculation of the Fisher information matrix can be completed in the center. In this way, the amount of communication and calculation required by the algorithm can be reduced.

\renewcommand{\algorithmicrequire}{\textbf{Server:}}  
\renewcommand{\algorithmicensure}{\textbf{Client}$(k, G)$\textbf{:}}
\begin{algorithm}[t]
  \caption{Federated Averaging (FedAvg)}
  \label{algorithm1}
  \begin{algorithmic}[1]
      \State $G$ is global model and $L$ is local model
      \Require
          \State initialize $G_{0}$
          \For {each round $t=1,2,3...$}
              \State $m \gets max(C \cdot K, 1)$\qquad //$K$ is number of clients and $C$ is selection ratio
              \State $S_{t} \gets$ random set of $m$ clients
              \For {each client $k \in S_{t}$}
                  \State $G_{t}^{k} \gets Client(k, G_{t-1})$
              \EndFor
              \State $G_{t} \gets \frac{1}{n_{S_{t}}} \sum_{k \in S_{t}} n_{k} \cdot G_{t}^{k}$ \qquad //$n$ is data number
          \EndFor 
      \Ensure
      \State $L^{k} \gets G$
        \For {each epoch}
          \For {each batch $(x, y)$ in client $k$}
              \State update $L^{k}$ to minimize $Loss(L(x), y)$
          \EndFor
        \EndFor
      \State return $L^{k}$ to server
  \end{algorithmic}
\end{algorithm}

\subsection{Temporal Ensembling}

Temporal Ensembling (TE) is a method of semi-supervised learning using consistency regularization, which is an imporved method based on $\rm \Pi$-model \cite{TE2017}. As the training target obtained in $\rm \Pi$-model is only based on the evaluation of the network obtained in last epoch, it can be expected to be noisy. TE alleviates this by aggregating the predictions of multiple previous network evaluations
into an ensemble prediction. After every training epoch, ensemble target $Z_{i}$ will be updated as follows:

\begin{equation}
  Z_{i} \gets \beta Z_{i} + (1-\beta) z_{i}
\end{equation}

where $z_{i}$ is the network output and $\beta$ is a momentum term. Experiments show that by using the ensemble target, the performance of the model will be significantly improved.

\section{New target in local training}

\subsection{Motivation}

The idea of many FL algorithms is to optimize FL by adding constraints to local training, such as FedProx \cite{FedProx}, FedCurv \cite{FedCurv} and FedCL \cite{FedProx}. These algorithms have one thing in common, that is, they all only use the model obtained in the last round of training as the constraint target in local training. However, we found that this choice is not necessarily good.

\begin{figure}
  \centering
  \includegraphics[height=180pt]{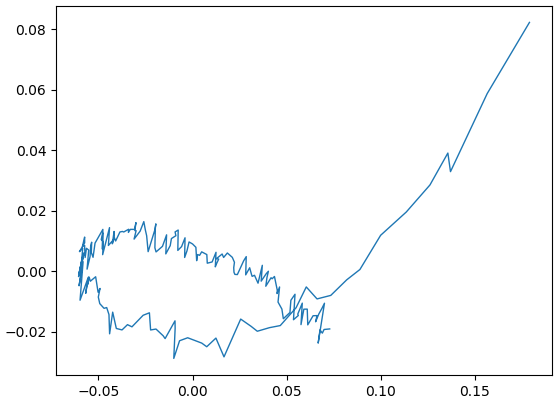}
  \caption{Changing trend of the global model}
  \label{fig1}
\end{figure}

In our research, we carefully studied the changing trend of the global model during the training process when the data distribution is non-iid. We use Principal Component Analysis (PCA) to reduce the dimension of the global model, and then connect it in time order to get Figure \ref{fig1}.

Through the figure, it can be found that the global model changes drastically during the training process. This shows that during the training process, due to the different data distributions of the nodes participating in each round of training, the model will have different biases compared to the optimal solution. Obviously, such a constraint target is not a good choice. Therefore, we need to design a new constraint target for this type of algorithm.

\subsection{Method}

The design of the new constraint target is mainly based on the following two points. The first is that we expect to get a stable constraint target. Secondly, we hope that this constraint target can provide guidance for the local training of the model.

At this point, we notice that our requirements match the TE scenario very well. Experiments of TE have proved that the ensemble prediction is a better prediction than the prediction of a single model, and it can help the model learn more knowledge. So we can apply the idea of TE to our algorithm. We use the aggregated results of all previous global models as the constraint target of local training. The details are shown in Algorithm \ref{algorithm2}, where $\beta$ is a momentum term and $T$ is the constraint target.

\renewcommand{\algorithmicrequire}{\textbf{Execute on Server:}}
\begin{algorithm}[H]
  \caption{Get Constraint Target}
  \label{algorithm2}
  \begin{algorithmic}[1]
      \Require
          \State $\hat T \gets 0$
          \For {each round $t=1,2,3...$}
              \State $\hat T \gets (1 - \beta) G_{t} + \beta \hat T$
              \State $T \gets \frac{\hat T}{1 - {\beta}^{t}}$
          \EndFor
  \end{algorithmic}
\end{algorithm}

\section{Experiments}

\subsection{Experimental Setup}

\subsubsection{Dataset}

We use three datasets for experiments: MNIST \cite{lecun1998gradient}, FashionMNIST \cite{xiao2017fashion} and CIFAR10 \cite{2009Learning}. In order to simulate the FL environment, we used Hsu's method \cite{HsuMeasuring} to divide the data set into different nodes. The data distribution of each node will be generated by the Dirichlet distribution as follows:

\begin{equation}
  \vec{q} \sim Dir(\gamma \vec{p})
\end{equation}

where $p$ is the global data distribution, $Dir$ is Dirichlet distribution 
and $\gamma$ is the parameter controlling the identicalness among clients. A bigger $\gamma$ indicates a more uniform distribution. For MNIST and FashionMNIST dataset we use $\gamma=1$ and for CIFAR10 dataset we use $\gamma=10$.We set up 10 clients and select 2 cilents for participating in training each round ($C=0.2$).

\subsubsection{Model}

We use Convolutional Neural Network (CNN) as the model in our experiments and the network architecture is shown in Table \ref{tabel1}.

\renewcommand{\arraystretch}{1.7}
\begin{table}[H]
  \centering
  \caption{Network Architecture}
  \label{tabel1}
  \begin{tabular}{|c|c|c|c|c|}
  \hline
  $\quad$Name$\quad$  & $\quad$Kernel size$\quad$ & $\quad$Stride$\quad$ & Output Channel & $\quad$Activation$\quad$ \\ \hline
  Conv1 & 5x5         & 1      & 16             & Relu       \\ \hline
  Pool1 & 2x2         & 2      & -              & -          \\ \hline
  Conv2 & 5x5         & 1      & 32             & Relu       \\ \hline
  Pool2 & 2x2         & 2      & -              & -          \\ \hline
  Fc   & -           & -      & 512            & Relu       \\ \hline
  Output   & -           & -      & 10             & Softmax    \\ \hline
  \end{tabular}
\end{table}

We set $E=2$ (local epoch) and $B=50$ (batch size) and use the Stochastic Gradient Descent (SGD) optimizer with learning rate 0.005 and a decay rate 0.99 per communication rounds.

\subsubsection{Evaluation}

In order to verify the effectiveness of the new constraint target, we apply it to FedProx and FedCL (FedCL and FedCurv are essentially the same, so we choose one of them), and the algorithm after replacing the constraint target is called FedProx-TE and FedCL-TE. The selection of hyper parameters in each experiment is shown in Table \ref{tabel2}. The choice of $\alpha$ is to make the algorithm perform best before replacing the constraint target, and the choice of $\beta$ is to make the modified algorithm perform best on the basis of a given $\alpha$.

\renewcommand{\arraystretch}{1.7}
\begin{table}[H]
  \centering
  \caption{Selection of hyper parameters}
  \label{tabel2}
  \begin{tabular}{|c|c|c|c|}
    \hline
                                  &              & $\qquad\alpha \qquad$ & $\qquad\beta\qquad$ \\ \hline
    \multirow{2}{*}{MNIST}        & $\quad$FedProx(-TE)$\quad$ & 1         & 0.2      \\ \cline{2-4} 
                                  & FedCL(-TE)   & 0.1       & 0.2      \\ \hline
    \multirow{2}{*}{$\quad$FashionMNIST$\quad$} & FedProx(-TE) & 1         & 0.2      \\ \cline{2-4} 
                                  & FedCL(-TE)   & 0.1       & 0.6      \\ \hline
    \multirow{2}{*}{CIFAR10}      & FedProx(-TE) & 0.4       & 0.4      \\ \cline{2-4} 
                                  & FedCL(-TE)   & 0.1       & 0.2      \\ \hline
    \end{tabular}
\end{table}

\subsection{Results and Analysis}

\begin{figure}[H]
  \centering
  \subfloat[MNIST FedProx(-TE)]{\includegraphics[width = .48\linewidth]{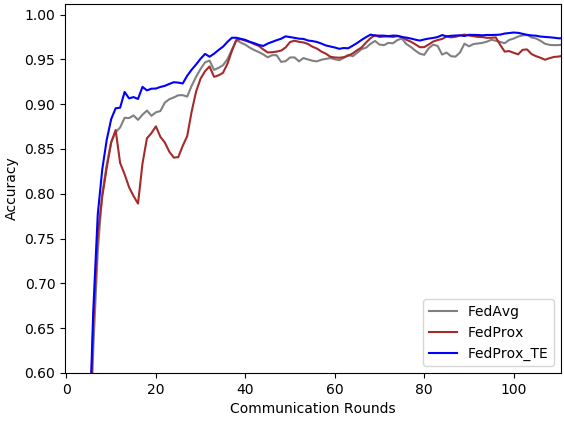}}
  \hfill
  \subfloat[MNIST FedCL(-TE)]{\includegraphics[width = .48\linewidth]{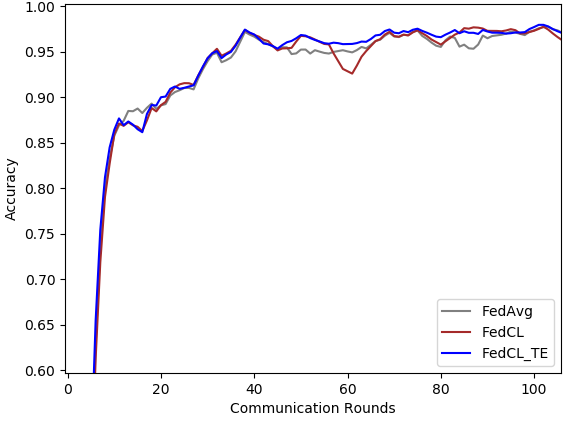}}
  \hfill
  \subfloat[FashionMNIST FedProx(-TE)]{\includegraphics[width = .48\linewidth]{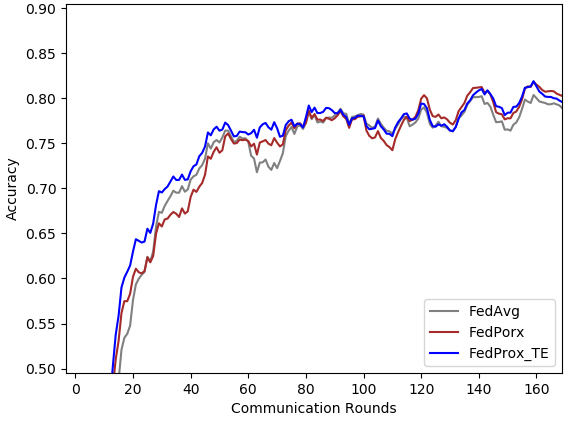}}
  \hfill
  \subfloat[FashionMNIST FedCL(-TE)]{\includegraphics[width = .48\linewidth]{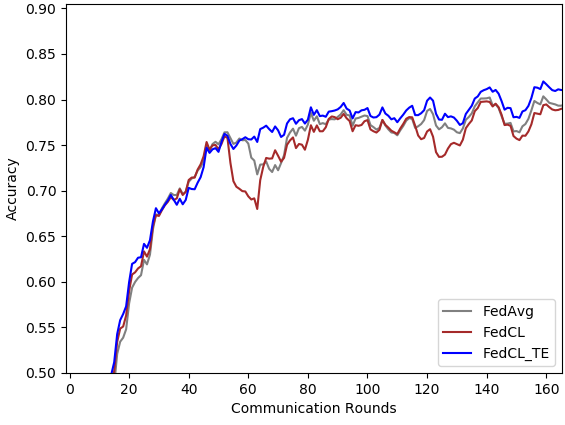}}
  \hfill
  \subfloat[CIFAR10 FedProx(-TE)]{\includegraphics[width = .48\linewidth]{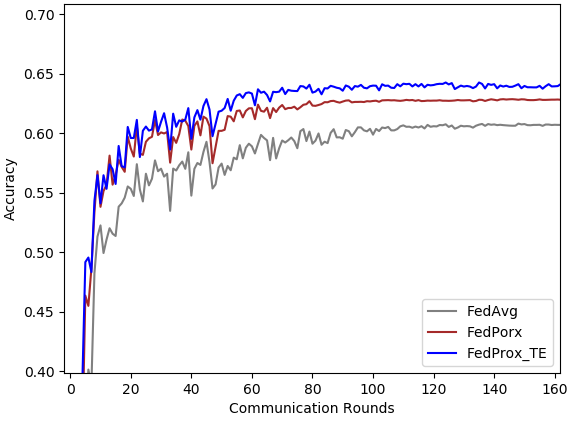}}
  \hfill
  \subfloat[CIFAR10 FedCL(-TE)]{\includegraphics[width = .48\linewidth]{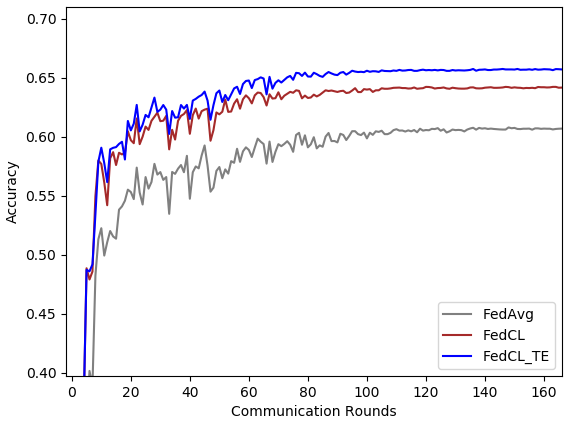}}
  \caption{Experimental results}
  \label{fig2}
\end{figure}

The experimental results are shown in Figure \ref{fig2}, (a)(b) are the results on MNIST dataset, (c)(d) are the results on FashionMNIST dataset and (e)(f) are the results on CIFAR10 dataset.

On the MNIST dataset, the number of communication rounds required for FedProx-TE to reach $95\%$ accuracy is reduced by $20\%$ compared to FedProx, but the new constraint target does not significantly improve the performance of FedCL. On the FashionMNIST dataset, the new constraint target significantly accelerates the training process of FedProx in the early stage. And compared with FedCL, the speed of FedCL-TE reaching $80\%$ accuracy has increased by $14\%$, and the accuracy of the model after convergence has increased by $2\%$. On the CIFAR10 dataset, the convergence speed of the two algorithms using the new constraint target has been significantly improved, and their accuracy has both increased by about $2\%$ after the model has converged.

In summary, the constraint target we designed can improve the performance of the original algorithm in most cases, and will not lead to a decrease in the performance of the algorithm in the worst case.

\section{Conclusion}

In this paper, we design a new constraint target for a class of FL algorithms that adds additional constraints in local training. We hope that the new constraint targets can provide better guidance for local training. Through experiments on FedProx and FedCL, we have proved that the new constraint target is effective. In the future work, we will try more methods to design the constrained target and apply it to more FL algorithms.

\end{document}